\documentclass[11pt]{article}

\usepackage[final]{acl}

\usepackage{times}
\usepackage{latexsym}

\usepackage[T1]{fontenc}
\usepackage{fontawesome}

\usepackage[utf8]{inputenc}

\usepackage{microtype}

\usepackage{inconsolata}

\usepackage{graphicx}

\usepackage{microtype}
\usepackage{hyperref}
\usepackage{url}
\usepackage{booktabs}
\usepackage{tcolorbox}
\usepackage{multirow}
\usepackage{caption}
\usepackage{subcaption}
\usepackage{soul}
\usepackage{todonotes}

\usepackage{amssymb}
\usepackage{pifont}
\newcommand{\cmark}{\ding{51}}%
\newcommand{\xmark}{\ding{55}}%

\title{Who Benchmarks the Benchmarks? Towards Comprehensive Evaluation of Commonsense Reasoning Benchmarks}

\author{
\noindent
\begin{minipage}[t]{0.2\textwidth}
    \centering
    ~~~\textbf{Pavel Chizhov}$^{1,2}$\\
\end{minipage}
\hfill
\begin{minipage}[t]{0.29\textwidth}
    \centering
    \textbf{Anton Changalidis}$^{2}$\\
\end{minipage}
\hfill
\begin{minipage}[t]{0.35\textwidth}
    \centering
    \textbf{Vishnu Prasad Vijaya Kumar}$^{1,2}$\\
\end{minipage}
\hfill
\begin{minipage}[t]{0.25\textwidth}
    \centering
    \textbf{Yannick Detrois}$^{2,3}$\\
\end{minipage}
\vspace{3mm} \\
\noindent
\begin{minipage}[t]{0.3\textwidth}
    \centering
    ~~~~~~~~~~~~~~~~~~~\textbf{Mattia Nee}$^{2}$\\
\end{minipage}
\hfill
\begin{minipage}[t]{0.33\textwidth}
    \centering
    \textbf{Pierre-Carl Langlais}$^{2}$\\
\end{minipage}
\begin{minipage}[t]{0.39\textwidth}
    \centering
    \textbf{Ivan P. Yamshchikov}$^{1,2}$~~~~~~~~~~~\\
\end{minipage}
\vspace{3mm} \\
$^{1}$CAIRO, Technical University of Applied Sciences Würzburg-Schweinfurt \\
$^{2}$PleIAs ~~~ $^{3}$EPFL\\
\small{
    \textbf{Correspondence:} \href{mailto:pavel.chizhov@thws.de}{pavel.chizhov@thws.de}
  }
}

\begin{document}
\maketitle

\begin{abstract}

Commonsense reasoning is a key language model capability, as it is purportedly a prerequisite for many basic tasks, unlike specific factual knowledge. It is often measured with multiple-choice questions (MCQ) benchmarks, e.g. HellaSwag and PIQA. Some of these benchmarks, however, are outdated and contain numerous validity issues. We illustrate some typical validity issues with a case study on HellaSwag, one of the most popular and problematic benchmarks for commonsense reasoning. 
The issues we find range from basic ungrammaticality and numerous typos to misleading prompts or equally correct options. We show that if we remove question prompts or replace them with \textit{"Lorem ipsum dolor..."}, about 68\% of model predictions do not change. We argue that this occurs due to inner flaws in the benchmark, not mere contamination that might be present in some models. 
Since benchmark scores are an essential part of model selection in both research and commercial applications, these issues can have severe consequences.
Based on our findings, we propose BenCheck, a package for benchmark validity analysis that encapsulates the main checks performed in our case study and can be used to audit commonsense reasoning benchmarks. We apply these checks to PIQA, Global PIQA, and Winogrande.
\begin{center}
\faGithub~\href{https://github.com/Pleias/BenCheck}{\path{PleIAs/BenCheck}}
    \end{center}
\end{abstract}

\section{Introduction} 
\label{sec:intro}

\begin{figure}
    \centering
    \includegraphics[width=\linewidth]{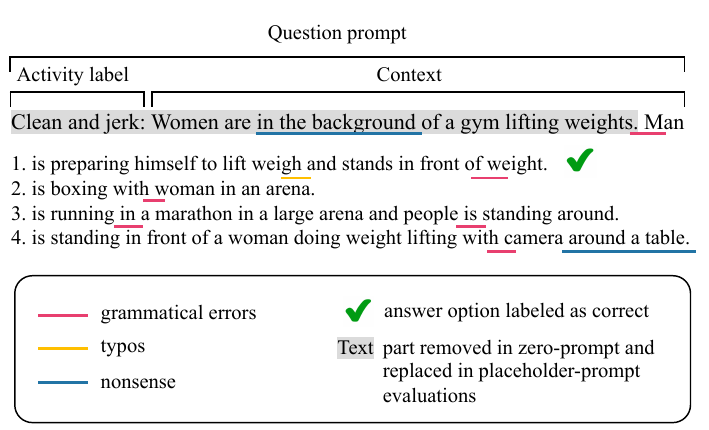}
    \caption{Example from the HellaSwag validation set. Validity issues are underlined in color. The part of the question prompt that we remove or replace in prompt-related experiments is highlighted in light gray.}
    \label{fig:example}
\end{figure}

Language models are evaluated on benchmarks. These evaluations shape model development by indicating how different design decisions, e.g., hyperparameters, training data, and post-training, impact model performance~\citep{biderman2024lessons}. NLP practitioners select training procedures to improve performance on accepted benchmarks. However, if the benchmarks do not measure what the researchers think they measure, development may not be going in the most optimal direction, and we may be missing out on performance gains.

\textbf{Construct validity} refers to whether a benchmark is measuring the capability it is claimed to measure \citep{cronbach1955construct,bean2025measuring}. 
Low construct validity can lead to real-world consequences, including misdirected research \citep{bean2025measuring}. Model developers choose dataset sources \citep{penedo2025fineweb} and make design choices (\citealp[][\textit{inter alia}]{palm2023chowdhery}) based on benchmark results. Furthermore, policymakers and government bodies worldwide may use benchmarks to inform policy decisions \citep{reuel2024betterbench}. 
Thus, benchmark construct validity not only influences model development and scientific research, but also even trickles down to policy-making processes. Without established validity, the use of a benchmark may be ``supercharging bad science'' \citep{blili2025stop}.

Commonsense reasoning is a desirable capability in a model because it encompasses a more generalizable language and world understanding rather than memorized facts.  
This kind of reasoning focuses on reasoning over domain-general scenarios and thus differs from the math- and code-oriented reasoning, for which there are increasingly many benchmarks. In a survey of over 100 commonsense reasoning benchmarks, \citet{davis2023benchmarks} observe widespread quality issues.
HellaSwag \citep{zellers-etal-2019-hellaswag} is one of the most widely used of these commonsense reasoning benchmarks, and it has played a large role in language model development over the past six years. 

In order for a commonsense reasoning benchmark to serve its purpose, it should allow us to draw an inference about model capabilities, specifically that the model with the highest HellaSwag score has the best commonsense reasoning capabilities. However, we argue that there are key issues (see an example in Figure~\ref{fig:example}) that prevent us from drawing that inference from  HellaSwag scores. We make the connection between dataset artifacts and construct validity as follows: the dataset artifacts (grammar/typos, generation hallucinations, etc.) are inherent flaws in the benchmark. The models exploit these artifacts and learn shortcuts that allow them to avoid the commonsense reasoning framework. Thus, the benchmark does not adequately measure the key target capability, thereby compromising construct validity. 

Our work makes the following contributions:

\begin{itemize}
    \item We conduct a case study on HellaSwag and argue that its scores are confounded by factors beyond sentence completion and commonsense reasoning. We point out diverse issues and their consequences and show that they are omnipresent in HellaSwag (Section~\ref{sec:annotations}). 
    \item We connect these issues to model performance and separately test HellaSwag's prompt importance by zero- and placeholder-prompt evaluations (Section~\ref{sec:model_evals}).
    \item Based on our case study, we introduce BenCheck (Section~\ref{sec:bencheck}), a package for dataset validity audits, and apply it to other commonsense reasoning benchmarks.
    \item We propose a list of requirements that a quality commonsense reasoning benchmark should meet (Section~\ref{sec:requirements}).
\end{itemize}

\section{Related Work}

\paragraph{Benchmark validity issues.}
Several recent studies have highlighted various issues in widely used benchmarks. 
In the text-generation evaluation format, MMLU \citep{hendrycks2021measuring} has been shown to yield variable performance for the same model after minor changes. 
Small perturbations to the prompt have also been shown to significantly affect model performance on benchmarks such as MMLU and HellaSwag~\citep{habba2025dove}.

In addition to sensitivity to these changes, MMLU has been shown to contain questions with no clear correct answer or incorrect ground truth labels \citep{gema2024we}. Other benchmarks, ranging in domain and task type, have also been shown to have errors in the reference answers or in the ground truth labels. One of these is GSM8K \citep{cobbe2021training}, a math word problem benchmark. \citet{vendrow2025large} find that at least 5\% of the items in GSM8K contain a serious error, such as mislabeled questions, logical contradictions, and ambiguous questions. Another example is XSUM \citep{narayan-etal-2018-dont}, a summarization benchmark. \citet{liang2023holistic} found that reference summaries in XSUM were rated as being worse than model-generated responses, according to human annotators. Therefore, the quality of the benchmark underestimates the model's summarization performance. 

The current paper contributes to this line of work, which aims to identify issues with existing benchmarks. We also encapsulate our analysis and other checks into a customizable package for benchmark validity audits, enabling faster, more comprehensive validation.

\paragraph{Commonsense reasoning benchmarks.}
Commonsense reasoning is a desirable capability in a language model because it requires relatively sophisticated language understanding and knowledge about the world. However, a survey of commonsense reasoning benchmarks highlights item quality as a common issue across over 100 text-based benchmarks \citep{davis2023benchmarks}, with HellaSwag among the most widely used.
A previously published blog post highlighted some issues with HellaSwag \citep{chen2023hellaswag}; however, the author used only a small sample ($n=300$) from the entire dataset. The author estimates that 36\% of items in HellaSwag contain errors. In this work, we annotate the entire validation set and expand the list of the issues we consider. As a result, we demonstrate that a much higher proportion of HellaSwag contains errors.  

\paragraph{Prompt importance.}
Prior work has shown that for many commonsense reasoning benchmarks, models may exploit task artifacts or option-level heuristics rather than the intended commonsense reasoning signal. \citet{trichelair-etal-2019-reasonable} show that in SWAG benchmarks, incorrect answer options differ from golden answers both structurally and in quality, and connect these artifacts with internal validity~\citep{campbell1963experimental} because they constitute an uncontrolled alternative explanation to the model’s performance. This, in turn, undermines the construct validity~\citep{cronbach1955construct} as the score variance does not reflect the target construct. \citet{kavumba-etal-2021-learning} fine-tune BERT and RoBERTa models on answer-only entries and show that models are able to learn shortcuts to distinguish correct answers without question context. Prompt-ablation diagnostics have also been used in ``no-context''~\citep{shah-etal-2020-expect}, ``answer-only''~\citep{li-etal-2022-systematic}, and ``choice-only''~\citep{balepur-rudinger-2024-large, balepur-etal-2024-artifacts} settings, where models can often complete the task without seeing the full prompt. 

Following this line of work, we perform zero-prompt and placeholder-prompt evaluations and analyze not only accuracy but also agreement and disagreement with the full-prompt predictions. This adds granularity to how prompt perturbations affect model performance, regardless of the actual score.

\paragraph{Benchmark validation.} Recent work introduced BenchMarker~\citep{balepur2026benchmarkereducationinspiredtoolkithighlighting}, a toolkit for LLM-as-a-judge evaluation of MCQ benchmarks, including the study of contamination. Our BenCheck complements it by covering checks that are largely inferred from the benchmark’s statistics and model behavior, with the two tools together providing a broader view of benchmark validity.

\section{Case Study: HellaSwag}
\label{sec:case_study}
HellaSwag is useful as a relatively large dataset, with over 10\,000 items in the validation set\footnote{The validation set is the only public split on Hugging Face (except for train) and is commonly used as the test set, which was intentionally hidden to avoid contamination.}. 
It is used in leaderboards, such as Hugging Face's Open LLM Leaderboard v1, which measure and encourage development of open-weight models. It also still often appears in the curated list of benchmarks for model cards and technical reports for production-scale language models, e.g. GPT-4 \citep{achiam2023gpt} and Qwen 2.5 \citep{yang2025qwen25}, and smaller open models, e.g. Gemma~3~\citep{google2024gemma3} and OLMo~3 \citep{olmo2025olmo}. Since its release, it has had an immeasurable impact on language model development and remains one of the most popular default evaluation choices.

HellaSwag was created by taking portions of existing data (ActivityNet~\citep{krishna2017dense} and WikiHow, which the authors scraped themselves) and generating alternative completions to the segments. The completions were generated by the original GPT model~\citep{radfordimproving}, largely inferior to the current state of the art, and curated using an adversarial filtering procedure intended to select only the machine-generated answers that seemed most plausible to the language model but nonsensical to humans. Human validators reviewed the dataset after the AF process. The task was intended to be trivial for people (approximately 95\% accuracy) but significantly more difficult (less than 50\% accuracy) for SOTA models at the time, which (1) were primarily encoder-style models and (2) are not necessarily relevant for current models, especially generative ones.
In this case study, we focus on the following issues:

\paragraph{Grammar errors and typos.} Grammatical errors can have opposite effects depending on where they occur: errors in the prompt or correct answer may make an item harder, whereas errors in distractors may make it easier by lowering the probability of those options. Although models may be expected to handle minor errors, robustness to such noise should be evaluated separately.

\paragraph{Nonsensicality.} Nonsensical text directly undermines commonsense reasoning evaluation, especially when it appears in the prompt or correct answer. If the distractors are implausible (see example in Section~\ref{sec:annotations}), the task becomes trivial and non-informative, just as with the grammar errors.

\paragraph{None or many good answers.} HellaSwag is designed to have exactly one correct answer. If no good option exists, models are rewarded for avoiding bad answers, which can overestimate performance. If multiple answers fit, a plausible choice may be marked wrong, which can underestimate performance and distort the random baseline.

\paragraph{Prompt importance.} As discussed in previous sections, commonsense reasoning is directly evaluated by inferring the answer option from the question prompt (context). If the context can be avoided and the answer is inferable from other heuristics, including the plausibility or grammatical correctness of the answer options, the construct validity of the benchmark is compromised.

\subsection{Annotations}
\label{sec:annotations}

We use Gemini 2.5 Flash~\citep{comanici2025gemini25pushingfrontier} to annotate the entire HellaSwag validation set. Based on the validity issues enumerated above, we formulate a list of rubrics for LLM annotation, see the full prompt in Appendix~\ref{sec:app-annotation}. We also specify in the prompt which answer is supposed to be correct so that we do not rely on LLM's solutions for the benchmark, and the model has a better understanding of the question's design. To check the reliability of LLM annotations, we collect two sets of human annotations for a subsample of the validation set. In most cases, LLM annotations received high agreement with at least one human annotator, and overall agreement was on a similar level as inter-annotator agreement. We present the validation analysis and provide further discussion in Appendix~\ref{sec:app-human}.

From the evaluations of grammar, we find that 30.9\% of question prompts, 24.1\% of correct answers, and the absolute majority of distractor sets (68.6\%) contain grammatical issues or typos (Table~\ref{tab:annotations}). For the ActivityNet part, this can be attributed to the benchmark creation methodology, in which these questions were generated by the original GPT model~\citep{radfordimproving}, which is drastically inferior to modern ones. It is also clear that correct answers generally have considerably fewer issues than incorrect ones because, on benchmark construction, the correct answers were taken from an existing corpus and the incorrect ones were synthetically generated. We also found that 85.1\% of questions have utterly implausible constructions in incorrect options, for example: 
\begin{quote}
    ``Changing out a car seat can dramatically improve the color of your car seat if you use \textbf{vinyl, vinyl, vinyl, or vinyl} seats.''
\end{quote} This might lead to trivial solutions dependent solely on choosing the least problematic option. 4.7\% of questions were also highlighted due to ethical concerns. We investigated that they usually contain unethical language expressions or descriptions of violence, but also describe making weapons, taking drugs, and adult content in several cases.

\begin{table}[!t]
        \small
        \centering
            \begin{tabular}{l c c c}
            \toprule
            Parameter & Total, \% & AN, \% & WH, \% \\
            \midrule
            Grammar/typos: & & & \\
            --- Prompt  &  30.9\% & 46.1\% & 23.6\% \\
            --- Correct option  &  24.1\% & 17.3\% & 27.3\% \\
            --- Incorrect option(s)  &  68.6\% & 54.6\% & 75.2\% \\
            Incoherent: & \\
            --- Prompt  &  4.0\% & 9.0\% & 1.6\% \\
            --- Correct option  &  0.8\% & 0.9\% & 0.8\% \\
            Illogical correct answer & 3.2\% & 4.2\% & 2.7\% \\
            Implausible distractors & 85.1\% & 67.8\% & 93.3\% \\
            \midrule
            Ethical concerns  &  4.7\% & 2.8\% & 5.5\% \\
            \midrule
            Wrong best answer & 1.1\% & 1.7\% & 0.9\% \\
            Multiple correct options  &  8.0\% & 11.7\% & 6.2\% \\
            \bottomrule
        \end{tabular}
        \caption{Gemini annotations for the HellaSwag validation set: for the complete set, and by source: from ActivityNet (AN) and WikiHow (WH). The validation set comprises 10042 questions: 3243 (32.3\%) from ActivityNet, and 6799 (67.7\%) from WikiHow.}
        \label{tab:annotations}
\end{table}

\begin{table}[htbp]
\centering
\small
\setlength{\tabcolsep}{4pt}
\begin{tabular}{lc}
\toprule
Accuracy
& 0.62 \\
\midrule
Prompt grammar/typos
& 0.59 \\
Incoherent prompt
& 0.54 \\
Correct option grammar/typos
& 0.52 \\
Incoherent correct option
& 0.38 \\
Illogical correct option
& 0.47 \\
Incorrect option(s) grammar/typos
& 0.65 \\
Implausible incorrect option(s)
& 0.64 \\

\bottomrule
\end{tabular}
\caption{Performance of Llama 3.2 1B on full HellaSwag validation set (top row) and on questions with highlighted issues. Full results reported in Appendix~\ref{sec:app-accuracy-vs-label}.}
\label{tab:accuracy_vs_label_llama}
\end{table}

\subsection{Model Evaluations}
\label{sec:model_evals}

\begin{figure*}[htbp]
\centering
    \includegraphics[width=\textwidth]{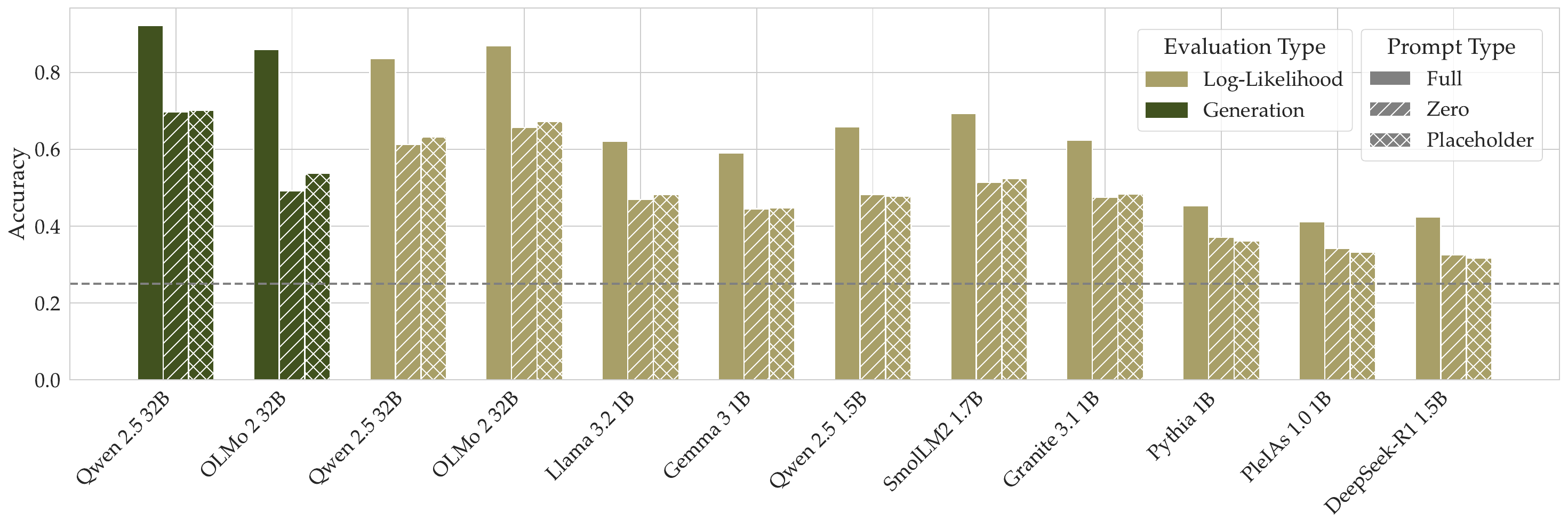}
    \caption{Accuracy evaluation with log-likelihood for larger (32B) and smaller (1-2B) models with a full question prompt,  and without it (zero-prompt), and with \textit{``Lorem ipsum''} instead of the prompt (placeholder-prompt).}
    \label{fig:accuracy}
\end{figure*}

\begin{figure}[t]
    \small
    \centering
        \includegraphics[width=0.9\linewidth]{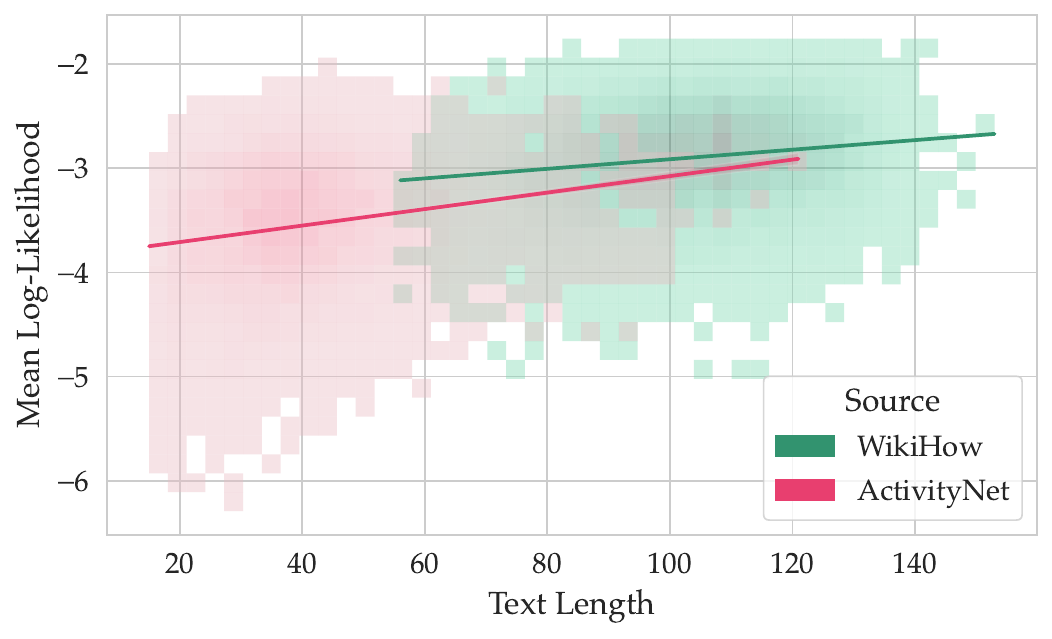}
        \caption{Mean log-likelihood vs answer option lengths. The lines represent the trend for each question source.}
        \label{fig:length_ll}
\end{figure}

We apply the preprocessing procedure from \texttt{lm-evaluation-harness}~\citep{biderman2024lessons} and perform two types of evaluations: log-likelihood and by generation. They differ in the way the task is presented. With log-likelihood, we directly request the model's estimate of the plausibility of each option. On the contrary, evaluation by generation allows us to present the model with all options, enabling it to compare and choose the most plausible or the least implausible option. We consider evaluation by generation only for larger models ($\sim$32B), as their instruct versions are more likely to produce outputs in the desired format. In addition, small model capabilities are often underestimated in prompt-based evaluation~\citep{hu-levy-2023-prompting} due to auxiliary task demands~\citep{hu2024auxiliary}.  We report the used models and the generation prompt in Appendices~\ref{sec:app-models}~and~\ref{sec:app-generation}. 

\textbf{Mean Log-Likelihood.~~} We append each answer option to the question prompt and run each sequence through the model to compute output logits. We then choose the answer with the highest mean log-likelihood (see formula in Appendix~\ref{app:log-likelihood}).

\textbf{Generation.~~} The model is presented with a question and all answer choices and is asked to output the correct answer label (1--4). The language model prompt used for evaluation is presented in Appendix~\ref{sec:app-generation}. We use the default generation strategy for each model. To lower the potential impact of contamination, we shuffle the options for each question, following \citet{alzahrani-etal-2024-benchmarks}.

\subsubsection{Accuracy evaluations}
\label{sec:issues-accuracy}

We first evaluate all selected models on the complete HellaSwag and separately on selected annotated categories related to common issues in prompts and answer options. We present the results in Table~\ref{tab:accuracy_vs_label_llama}. As hypothesized above, issues with question prompts and correct answer options make the task more challenging, without increasing the question's value in terms of commonsense reasoning. When the incorrect options are ungrammatical or contain utterly nonsensical expressions, the task becomes simpler. All tested models exhibit similar trends (see Appendix~\ref{sec:app-accuracy-vs-label}).
We also find that questions from WikiHow are generally simpler for all models (see details in Appendix~\ref{sec:app-accuracy-source}).

\begin{figure}[tbp]
\centering
    \includegraphics[width=0.9\linewidth]{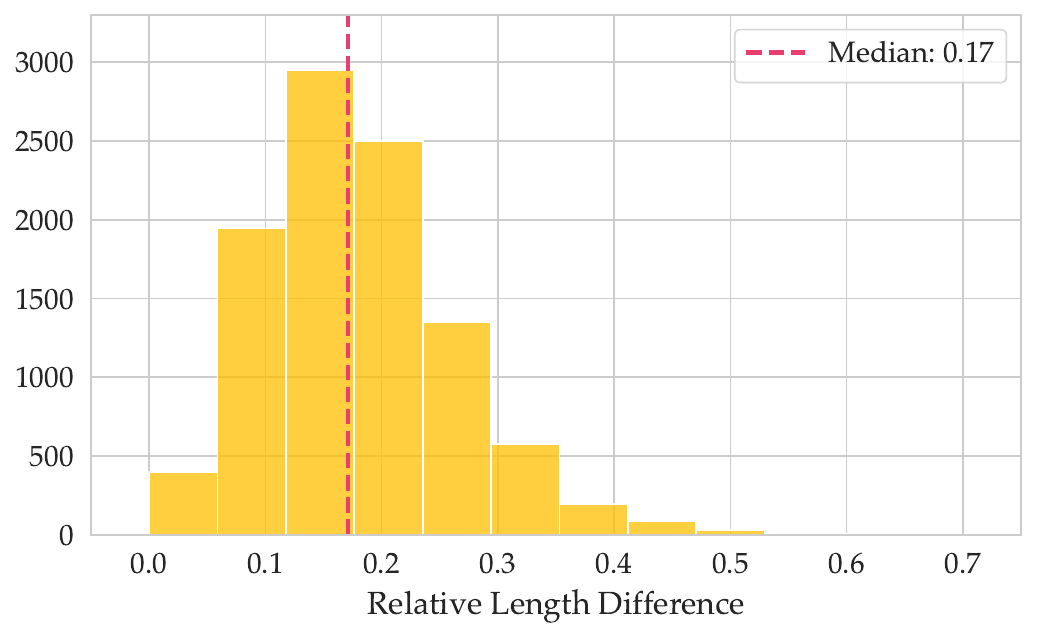}
    \caption{Differences between the longest and the shortest answer options relative to the longest option length.}
    \label{fig:length_difference}
\end{figure}

\begin{table*}[!t]
\centering
\small
\begin{tabular}{llccccccc}
\toprule
\multirow{2.4}{*}{Evaluation} & \multirow{2.4}{*}{Model} & \multirow{2.4}{*}{Params} & \multirow{2.4}{*}{Agreement} & \multicolumn{2}{c}{Agreement type} & \multicolumn{3}{c}{Disagreement type} \\ 
\cmidrule(lr){5-6} \cmidrule(rl){7-9}
 &  &  &  & Both \cmark & Both \xmark  & Full \cmark & Zero \cmark & Both \xmark \\
 \midrule
    \multirow{10}{*}{Log-Likelihood} & Llama 3.2 & \hphantom{1.}1B & 0.69 & 0.43 & 0.26 & 0.19 & 0.04 & 0.07 \\
    & Gemma 3 & \hphantom{1.}1B & 0.70 & 0.40 & 0.29 & 0.19 & 0.04 & 0.07 \\
    & Qwen 2.5 & 1.5B & 0.69 & 0.45 & 0.24 & 0.21 & 0.03 & 0.07 \\
    & SmolLM2 & 1.7B & 0.71 & 0.49 & 0.23 & 0.21 & 0.03 & 0.05 \\
    & Granite 3.1 & \hphantom{1.}1B & 0.72 & 0.44 & 0.28 & 0.18 & 0.03 & 0.06 \\
    & Pythia & \hphantom{1.}1B & 0.71 & 0.32 & 0.39 & 0.14 & 0.05 & 0.10 \\
    & PleIAs & \hphantom{1.}1B & 0.69 & 0.28 & 0.41 & 0.13 & 0.06 & 0.11 \\
    & DeepSeek-R1 & 1.5B & 0.65 & 0.26 & 0.39 & 0.16 & 0.06 & 0.13 \\
    & Qwen 2.5 & 32B & 0.71 & 0.60 & 0.12 & 0.24 & 0.02 & 0.03 \\
    & OLMo 2 & 32B & 0.74 & 0.64 & 0.09 & 0.23 & 0.01 & 0.02 \\
    \midrule
    \multirow{2}{*}{Generation} & Qwen 2.5 & \hphantom{.}32B & 0.70 & 0.67 & 0.03 & 0.25 & 0.03 & 0.02 \\
    & OLMo 2 & \hphantom{.}32B & 0.50 & 0.45 & 0.05 & 0.41 & 0.05 & 0.04 \\
 \bottomrule
\end{tabular}
\caption{The breakdown of zero-prompt agreement (a model gives equal correct or incorrect predictions) and disagreement (the model gives correct prediction only in either full- or zero-prompt scenario or gives different incorrect answers).}
\label{tab:agreement}
\end{table*}

\begin{table*}[htbp]
\small
\centering
\begin{tabular}{lcccccccccc}
\toprule
Number of models & 1 & 2 & 3 & 4 & 5 & 6 & 7 & 8 & 9 & 10 \\
 \midrule
    \% of questions & 75\% & 67\% & 60\% & 55\% & 50\% & 45\% & 39\% & 34\% & 27\% & 18\% \\
 \bottomrule
\end{tabular}
\caption{Zero-prompt core. A pair (N models, X\%) means that at least N models can answer correctly X\% of questions without question prompt texts.}
\label{tab:zero-prompt-core}
\end{table*}

\subsubsection{Answer length}

The likelihood of the answers positively correlates with their lengths in number of tokens (see Figure~\ref{fig:length_ll}), especially for the questions from ActivityNet. This might be due to the difference in answer option sources: generated distractor options are often shorter than the correct answers. Additionally, every text has very low likelihood values in the beginning. As the prior context grows longer, the following tokens become more predictable, and log-likelihoods increase overall. Therefore, the overall mean token probability is lower for shorter texts. This relationship may also be sensitive to the type of log-likelihood normalization. We separately investigate total and byte-normalized log-likelihoods proposed by~\citet{eleutherai2022multiple} and find that these methods also produce values correlated with the input length (see Appendix~\ref{app:log-likelihood} for details).

For HellaSwag questions, this issue is highly relevant. As we show in Figure~\ref{fig:length_difference}, the relative length difference between answer options, computed as the difference between the longest and the shortest answer options relative to the length of the longest answer option, can be quite high (more than 17\% for half of the data). Furthermore, we find that the accuracy of Llama 3.2 1B when the correct answer is the longest one (0.72) is greater than when it is not (0.59). Thus, varied answer lengths pose validity problems for a benchmark, as models implicitly prefer longer answers.

\subsubsection{Prompt importance}
\label{sec:zero-prompt}

We test the construct validity of HellaSwag in zero- and placeholder-prompt evaluations. In the \textbf{zero-prompt} scenario, we remove the activity label and the question context from the task; in the \textbf{placeholder-prompt}, we substitute them with a generic text typically used as a filler:

\begin{center}
    \parbox{0.8\linewidth}{\textit{Lorem ipsum dolor sit amet, consectetur adipiscing elit. Morbi vel venenatis dui. Pellentesque sed cursus massa.}}
\end{center}

\noindent We show an example of the affected part of the question prompt in Figure~\ref{fig:example}. In commonsense reasoning, the model should figure out which ending stems best from the question prompt. By comparing the results from these evaluations, we can determine whether this has a causal impact on model performance. If a model can choose the correct option without the question text, this compromises the construct validity.

\begin{table*}[!t]
\small
\centering
\begin{tabular}{llccccccc}
\toprule
\multirow{2.4}{*}{Evaluation} & \multirow{2.4}{*}{Model} & \multirow{2.4}{*}{Size} & \multirow{2.4}{*}{Agreement} & \multicolumn{2}{c}{Agreement type} & \multicolumn{3}{c}{Disagreement type} \\ 
\cmidrule(lr){5-6} \cmidrule(rl){7-9}
& &  &  & Both \cmark & Both \xmark  & Full \cmark & Lorem \cmark & Both \xmark \\
 \midrule
    \multirow{10}{*}{Log-Likelihood} & Llama-3.2 & 1B & 0.67 & 0.43 & 0.25 & 0.20 & 0.06 & 0.08 \\
     & Gemma-3 & 1B & 0.69 & 0.40 & 0.29 & 0.19 & 0.05 & 0.07 \\
     & Qwen-2.5 & 1.5B & 0.65 & 0.43 & 0.22 & 0.23 & 0.05 & 0.07 \\
     & SmolLM2 & 1.7B & 0.64 & 0.46 & 0.18 & 0.23 & 0.06 & 0.07 \\
     & Granite-3.1 & 1B & 0.71 & 0.44 & 0.27 & 0.18 & 0.04 & 0.07 \\
     & Pythia & 1B & 0.55 & 0.26 & 0.30 & 0.19 & 0.10 & 0.15 \\
     & PleIAs & 1B & 0.59 & 0.24 & 0.35 & 0.17 & 0.09 & 0.14 \\
     & DeepSeek-R1 & 1.5B & 0.62 & 0.25 & 0.37 & 0.18 & 0.07 & 0.14 \\
     & Qwen-2.5 & 32B & 0.71 & 0.61 & 0.10 & 0.23 & 0.02 & 0.03 \\
     & OLMo-2 & 32B & 0.73 & 0.65 & 0.08 & 0.22 & 0.02 & 0.02 \\
     \midrule
    \multirow{2}{*}{Generation} & Qwen-2.5 & 32B & 0.70 & 0.67 & 0.03 & 0.25 & 0.03 & 0.02 \\
     & OLMo-2 & 32B & 0.54 & 0.49 & 0.05 & 0.37 & 0.05 & 0.04 \\
 \bottomrule
\end{tabular}
\caption{Agreement between full-prompt and placeholder-prompt evaluation. Agreement means that a model gives equal predictions in both evaluation modes. We separately report fractions for agreement types (a model gives equal correct or incorrect predictions) and disagreement types (the model gives correct prediction only in either full- or placeholder-prompt scenario or gives different incorrect answers).}
\label{tab:agreement-lorem}
\end{table*}

\paragraph{Accuracy.} For all models (see Figure~\ref{fig:accuracy}), accuracies with both prompt perturbations are well above chance (25\%), meaning the probabilities of the answer choices are biased towards the correct answer. This presents two problems. First, the question prompt does not contribute to the task. The model needs only to choose the most plausible text fragment, which does not necessarily reflect commonsense reasoning capabilities. Second, features of the incorrect answers, such as grammatical or logical errors, as discussed in Section~\ref{sec:annotations}, may increase the probability of the correct option. A model may therefore achieve high accuracy on a subset of the questions by simply ruling out grammatical inconsistencies or nonsensical phrases.

\paragraph{Agreement.} We separately test the agreement of predictions with full and modified prompts. Following~\citet{lyu-etal-2024-beyond}, ``agreement'' here means that the model gives the same predictions in both evaluation modes (with and without the prompt), regardless of whether these predictions are correct or wrong. The results of agreement evaluations are presented in Table~\ref{tab:agreement} for all models. On average, 68\% of the model predictions do not change if we remove the question prompt from the evaluation. This holds not only for correct answers but also for a substantial proportion of the incorrect ones, which rules out contamination as the main suspected cause. For Pythia, PleIAs, and DeepSeek-R1 models, the share of agreement for incorrect answers is larger than for the correct ones. This suggests that, for the majority of HellaSwag questions, the question text does not contain additional information that the model uses to perform the task.
Furthermore, for some questions, removing the context actually allowed the model to make a better choice (see column Zero \cmark).
We observe similar agreement patterns in the placeholder-prompt scenario (see Table~\ref{tab:agreement-lorem}), with even more questions (up to 10\%) answered correctly with a perturbed prompt.

In addition, we find that questions answered correctly without the prompt are shared across many models. In Table~\ref{tab:zero-prompt-core}, we show the percentage of questions answered correctly without the prompt by groups of models. In particular, about 18\% of questions are answered correctly without prompts by all 10 tested models. Together with the other results, these findings lead us to question the most fundamental quality of the benchmark --- its construct validity. Since the question text does not play a decisive role in $\sim$68\% of cases, the benchmark cannot accurately measure commonsense reasoning, as presenting the model with the cause does not help it choose the right effect. This also raises concerns about the answer choices: the incorrect options may often be so blatantly implausible that the correct answer stands out, even without context.

\begin{figure}
    \centering
    \includegraphics[width=\linewidth]{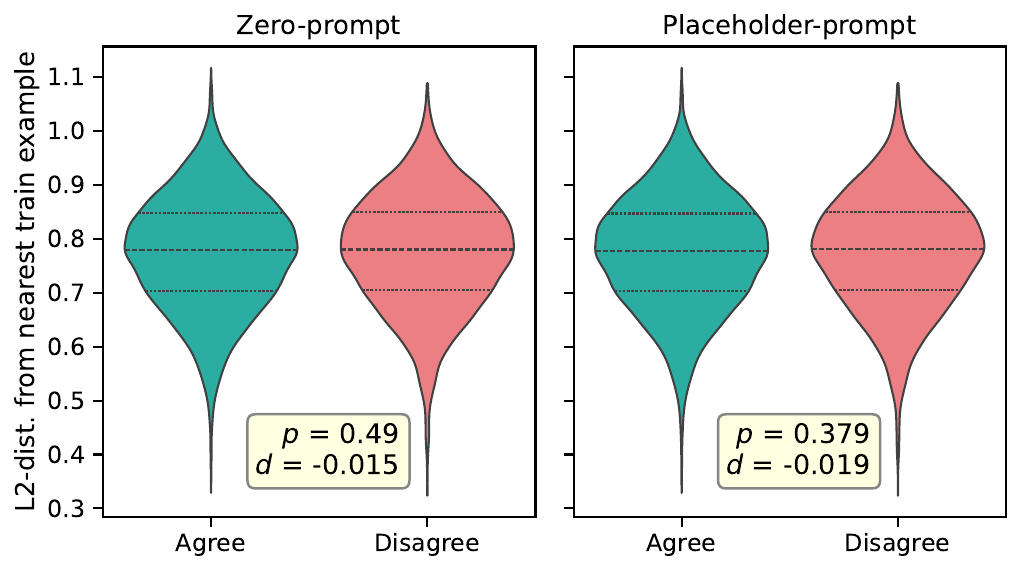}
    \caption{Distribution of L2 distances to near-duplicates in the train set for zero-/placeholder-prompt agreement for Llama 3.2 1B. For each pair, we present the p-value of the t-test and Cohen's d (both insignificant).}
    \label{fig:dist-to-train-llama}
\end{figure}

Finally, we test how zero-/placeholder-prompt agreement corresponds to the distance to the nearest neighbor in the training set (see results for Llama 3.1 1B in Figure~\ref{fig:dist-to-train-llama} and Appendix~\ref{app:dist-to-train} for other models). We use the \texttt{all-MiniLM-L6-v2} sentence transformer~\citep{reimers-2019-sentence-bert} and L2 distance. Both comparisons show no significant difference, indicating that if HellaSwag was present in the Llama's training set, it has no effect on the zero-/placeholder-prompt agreement, ruling out contamination as a primary alternative explanation. This pattern shows in other models, regardless of their openness and performance (Appendix~\ref{app:dist-to-train}).

\subsection{Interim Summary}

We identify dimensions relevant to construct validity and annotate the questions. We find that questions with issues in prompts or correct answers are harder for the models, whereas those with issues in the distractors make the task simpler. We also see that the question prompt does not play the decisive role in the majority of questions, with 18\% of questions being answerable without the prompt for all ten tested models. However, this analysis is limited to a single benchmark. Scaling this analysis to other benchmarks is time-intensive; therefore, in the following section, we propose a tool to quickly identify potential validity issues in commonsense reasoning benchmarks.

\section{BenCheck Package}
\label{sec:bencheck}

Based on our results and related work, we create and release a package for benchmark validity analysis under a permissive license --- \textbf{BenCheck}\footnote{\url{https://github.com/Pleias/BenCheck}}. In this package, we combine the main tests from our case study with several other relevant works on benchmark validation. Each of the implemented checks is accompanied by diagnostics and a detailed report with metric explanation. In particular, we include the following customizable checks:
\begin{itemize}
    \item \textbf{Answer lengths}: we compare answer lengths of correct and incorrect answer options, relative answer length difference, and also compute the proportion of cases when the correct answer is the longest or the shortest one.
    \item \textbf{Enumeration bias}: following the work of~\citet{alzahrani-etal-2024-benchmarks}, we check the distribution of correct answers among option positions as well as model preferences or biases.
    \item \textbf{Prompt importance}: following our evaluations with zero-prompt and placeholder-prompt (Section~\ref{sec:zero-prompt}), we compare the scores with the original and removed or replaced question text, thereby evaluating the question prompt importance and construct validity of the benchmark.
    \item \textbf{Placeholder answer}: we replace the correct answer option with ''none of the above'' (NOTA), as in the work of~\citet{salido2025othersgeneraltechniquedistinguish}, and measure changes in model performance. For example, if a model changes from an incorrect answer to 'NOTA', the original correct answer may present issues.
    \item \textbf{Grammar and typos}: We provide an interface for any open LLM with guided decoding or optionally GECToR~\cite{omelianchuk-etal-2020-gector}, a grammar correction model, for detecting grammatical errors and typos in question prompts, correct and incorrect answer options.
\end{itemize}

In Table~\ref{tab:bencheck}, we present main metrics from BenCheck reports for commonsense reasoning benchmarks: Winogrande~\citep{sakaguchi2019winogrande}, PIQA~\citep{Bisk2020}, and the English subset from Global PIQA~\cite{chang2025globalpiqaevaluatingphysical}, a manually corrected, multilingual version of the dataset. 

Winogrande exhibits the most grammatical errors in question prompts, with no signs of grammatical issues in answer options, given its single-word answers. For the same reason, it is the only benchmark robust to zero-prompt and placeholder-prompt evaluations. Compared to PIQA, Global PIQA presents fewer grammatical issues, which confirms better dataset curation, while still being solvable without the question prompts. Similarly to Section~\ref{sec:issues-accuracy}, we present additional model evaluations based on grammar issues present in benchmark parts in Appendix~\ref{app:grammar-issues}. HellaSwag's case of varied answer lengths is absent in other benchmarks; in terms of position bias, all benchmarks have uniformly distributed options. In a NOTA scenario, the task becomes more complicated, and the models rarely choose the correct answer, which might also indicate that, in some cases, the distractors are suitable. At the same time, in 24-32\% of cases, the model chooses NOTA instead of a previously incorrect answer, which indicates that the correct answer might present issues.

\begin{table}[!t]
\centering
\scriptsize
\begin{tabular}{lcccc}
\toprule
\textbf{Metric} & \textbf{HellaSwag} & \textbf{Winogrande} & \textbf{PIQA} & \textbf{G-PIQA} \\
\midrule
\multicolumn{5}{c}{\textbf{Prompt importance}} \\
\midrule
Accuracy & 0.792 & 0.589 & 0.864 & 0.846 \\
Zero score & \textcolor{red}{0.646} & 0.498 & \textcolor{red}{0.771} & \textcolor{red}{0.797} \\
\hphantom{---}Agreement & \textcolor{red}{0.588} & 0.496 & \textcolor{red}{0.813} & \textcolor{red}{0.770} \\
Lorem score & \textcolor{red}{0.523} & 0.490 & \textcolor{red}{0.714} & \textcolor{red}{0.700} \\
\hphantom{---}Agreement & \textcolor{red}{0.564} & 0.510 & \textcolor{red}{0.800} & \textcolor{red}{0.700} \\
\midrule
\multicolumn{5}{c}{\textbf{Grammar issues}} \\
\midrule
Question & \textcolor{red}{0.35} & \textcolor{red}{0.39} & \textcolor{red}{0.20} & 0.03  \\
Golden & \textcolor{red}{0.12} & 0.00 & \textcolor{red}{0.13} & \textcolor{red}{0.12} \\
Distractors & \textcolor{red}{0.60} & 0.00 & \textcolor{red}{0.34} & \textcolor{red}{0.15} \\
\midrule
\multicolumn{5}{c}{\textbf{Answer lengths}} \\
\midrule
Mean rel diff & \textcolor{red}{0.43} & 0.01 & 0.08 & 0.02 \\
\cmark~is longest & 0.23 & 0.12 & 0.24 & 0.21 \\
\cmark~is shortest & 0.23 & 0.13 & 0.30 & 0.16 \\
\midrule
\multicolumn{5}{c}{\textbf{Enumeration bias}} \\
\midrule
A \cmark & 2505 & 628 & 910 & 50 \\
B \cmark & 2517 & 639 & 928 & 50 \\
C \cmark & 2552 & --- & --- & --- \\
D \cmark & 2468 & --- & --- & --- \\
\midrule
\multicolumn{5}{c}{\textbf{None of the above (NOTA)}} \\
\midrule
\cmark$\rightarrow$ NOTA & 0.06 & 0.05$^*$ & 0.08$^*$ & 0.06$^*$ \\
\xmark~$\rightarrow$ NOTA & \textcolor{orange}{0.32} & \textcolor{orange}{0.29}$^*$ & \textcolor{orange}{0.24}$^*$ & \textcolor{orange}{0.28}$^*$ \\
\bottomrule
\end{tabular}
\caption{BenCheck evaluation of commonsense reasoning benchmarks. We perform model evaluations with Gemma 3 12B~\citep{google2024gemma3} and grammatical issue detection with Gemma 4 E4B~\cite{gemma4_hf_blog_2026}. *The benchmark questions have only two options.}
\label{tab:bencheck}
\end{table}

\section{Discussion}
\label{sec:discussion}

Our experiments highlight a variety of artifacts in HellaSwag. They range from poor language to flaws in question design. One of the primary drivers of these issues is the outdated generation process. ActivityNet and WikiHow, the two subsets of HellaSwag, show different patterns. ActivityNet has generally shorter answer options and more grammatical issues. We hypothesize that this is due to the original source of the questions, the generated passages that describe the video content.

Even though the benchmark is generally considered challenging, this is mainly due to grammatical errors, typos, illogical and implausible constructions in many questions. Even though such difficulties can also test a model's capabilities, they are orthogonal to commonsense reasoning. Thus, under our diagnostics, HellaSwag is a weak proxy for commonsense reasoning, as it is not possible to isolate this capability judging solely by HellaSwag scores. Although implausible constructions allow the model to rule out incorrect options, effectively mimicking how humans write tests, we argue that this should occur only in connection with question prompts, not within answer options in isolation. 

Nevertheless, given that the intended correct option is still the most plausible (or the least problematic) in the absolute majority of items, the benchmark still presents at least some useful signal for model capabilities; these capabilities, however, are heavily conflated, and HellaSwag alone cannot be used for selecting the best model, especially as a proxy for commonsense reasoning specifically.

We encapsulate our findings, along with other relevant work, into a package for benchmark validity audits, a useful tool for screening and identifying critical issues. BenCheck checks can also be applied to MCQ benchmarks in general. In some cases, BenCheck can also be used to filter out faulty items from the benchmark. However, it must be ensured that the remaining items are enough for a comprehensive evaluation. For instance, filtering out problematic items from HellaSwag would remove the absolute majority of ActivityNet, effectively reducing the benchmark to tutorial-like items from WikiHow.

\section{Requirements.} 
\label{sec:requirements}

Based on our findings, we formulate a list of requirements that a valid modern commonsense reasoning benchmark should meet:

\begin{itemize}
\item \textbf{Grammar and typos.~~} All the texts should be grammatically correct, including the incorrect options. If the model is to be tested for robustness to bad grammar or typos, a separate version of the dataset should be created, e.g., by injecting errors or typos.

\item \textbf{Sensicality and prompt importance.~~} All options should be a reasonable, coherent text. Even though, by the nature of the task, incorrect options make no sense, this should be only due to how they relate to the question context and not, for instance, due to single implausible constructions. Apart from this, the question prompt should play a role in determining the correct answer.

\item \textbf{Distinct correct answers.~~} The correct option should be a substantially better fit for the question prompt than the incorrect options. There might be more than one correct option if the task design allows it, but they should all be comparably suitable.

\item \textbf{Uniform option lengths.~~} The answers can have different lengths, though this variability should be limited. The length rank of correct answers should be uniform over the dataset. The intended evaluation procedure should consider the potential length bias.

\item \textbf{Content.~~} The questions should be filtered for toxicity and ethical issues. For smaller models, the questions should preferably be on general-domain topics, excluding overly specific professional knowledge.
\end{itemize}

\section{Conclusion}

In this paper, we describe numerous issues in HellaSwag, ranging from grammatical errors and typos to ambiguous answer choices, and argue that these issues compromise the construct validity of the benchmark. Evaluations with perturbed prompts show that in most of the HellaSwag questions, question text does not affect model predictions, which allows us to conclude that HellaSwag performance does not necessarily indicate commonsense reasoning capabilities. 
We specifically highlight that commonsense reasoning is a salient characteristic, and there should be a raised research interest in constructing a good benchmark to measure it. We extend our findings to other benchmarks and propose an extensive package for benchmark validity analysis, providing an evaluation toolkit that guides the development of benchmarks.

\section*{Limitations}

The proposed methodology is diverse and requires different tools and different hardware for different checks. Model-related checks require models that can complete the task above a random chance and follow the instructions, as in the case with the generative evaluation; they also require GPU resources. Some of the checks are not necessarily indicative in all cases. For example, length or grammar checks are not representative of benchmarks with factual answers or the answers consisting of a single word. Additionally, evaluations of grammar relying on language models have their limitations. Customized grammatical error detection with specialized models is possible (e.g., as with GECToR), but the benchmark format should be carefully tailored for such evaluations.

The interpretation of some evaluations involves multiple possible explanations due to potential contamination of the model by the benchmark questions or content. We argue that our results show this behavior is present even without contamination (e.g., in models trained on open datasets); however, we cannot completely rule out this issue in closed-data models.

BenCheck is a useful tool for initial smoke testing of the benchmark. However, it is more of an audit toolkit that provides the aggregated picture rather than a precise filtering instrument. BenCheck-guided filtering is possible but has limitations, primarily in the choice of relevant criteria and the potential specificity of the remaining data. For instance, if HellaSwag were to be BenCheck-filtered, most ActivityNet examples would be removed, leaving only a subset of WikiHow. Though it is likely of higher quality, this dataset would represent only a narrow set of specifically formatted tutorials, which does not tell much about the model's capabilities.

Finally, though it has multiple issues and compromised construct validity, HellaSwag may still provide a useful engineering signal. The models with HellaSwag performance above random show better capabilities, which, when combined with other benchmarks on the leaderboards, might yield useful rankings in certain cases.

\section*{Acknowledgments}
The authors of this work acknowledge the HPC resource allocation by Erlangen National High-Performance Computing Center (NHR@FAU) of the Friedrich-Alexander-Universität Erlangen-Nürnberg (FAU) (joint project with the Center for Artificial Intelligence (CAIRO), THWS) and Jean Zay (compute grant \#GC011015451).
The authors would also like to thank Catherine Arnett for her thoughtful ideas and continuous help throughout this work, and Andrey Indukaev for manual annotations and useful ideas.

\bibliography{custom}

\appendix

\section{Models}
\label{sec:app-models}

\begin{table*}[htbp]
\small
\centering
\begin{tabular}{@{}lcl@{}}
\toprule
Model & Size & HuggingFace repository \\
 \midrule
LLaMA 3.2 & \multirow{2}{*}{1B} & \multirow{2}{*}{\texttt{meta-llama/Llama-3.2-1B}} \\
~~~~~~~~\citep{grattafiori2024llama3herdmodels} & & \\
Qwen 2.5 \citep{yang2025qwen25} & 1.5B & \texttt{Qwen/Qwen2.5-1.5B} \\
Granite 3.1 \citep{ibm2024granite31} & 1B & \texttt{ibm-granite/granite-3.1-1b-a400m-base} \\
Gemma 3 \citep{google2024gemma3} & 1B & \texttt{google/gemma-3-1b-pt} \\
SmolLM2 \citep{allal2025smollm2smolgoesbig} & 1.7B & \texttt{HuggingFaceTB/SmolLM2-1.7B} \\
DeepSeek-R1 \citep{deepseekai2025deepseekr1incentivizingreasoningcapability} & 1.5B & \texttt{deepseek-ai/DeepSeek-R1-Distill-Qwen-1.5B} \\
Pythia \citep{biderman2023pythia} & 1B & \texttt{EleutherAI/pythia-1b} \\
Pleias 1.0 \citep{LANGLAIS2025146} & 1B & \texttt{PleIAs/Pleias-1b-Preview} \\
\midrule
Qwen 2.5 \citep{yang2025qwen25} & 32B & \texttt{Qwen/Qwen2.5-32B-Instruct} \\
OLMo 2 \citep{olmo20252olmo2furious} & 32B & \texttt{allenai/OLMo-2-0325-32B-Instruct} \\
 \bottomrule
\end{tabular}
\caption{Model names and their corresponding HuggingFace repositories.}
\label{tab:models}
\end{table*}

In Table~\ref{tab:models}, we present the models we used for evaluations in the paper with their corresponding HuggingFace repository paths.

\section{Generation Prompt}
\label{sec:app-generation}

We use the following prompt to evaluate the models by generation:

\begin{tcolorbox}[
    title = \textbf{Generation Prompt}, 
]

You are given a situation followed by four possible endings. Choose the most appropriate ending by selecting the corresponding number. Respond only with the number of the correct answer. 

\vspace{0.5em}

\textbf{Context:} Roof shingle removal: A man is sitting on a roof. He

1. is using wrap to wrap a pair of skis.

2. is ripping level tiles off.

3. is holding a Rubik’s cube.

4. starts pulling up roofing on a roof.

\vspace{0.5em}

\textbf{Answer:}

\end{tcolorbox}

\section{Accuracy by Source}
\label{sec:app-accuracy-source}

\begin{figure*}
    \centering
    \includegraphics[width=\linewidth]{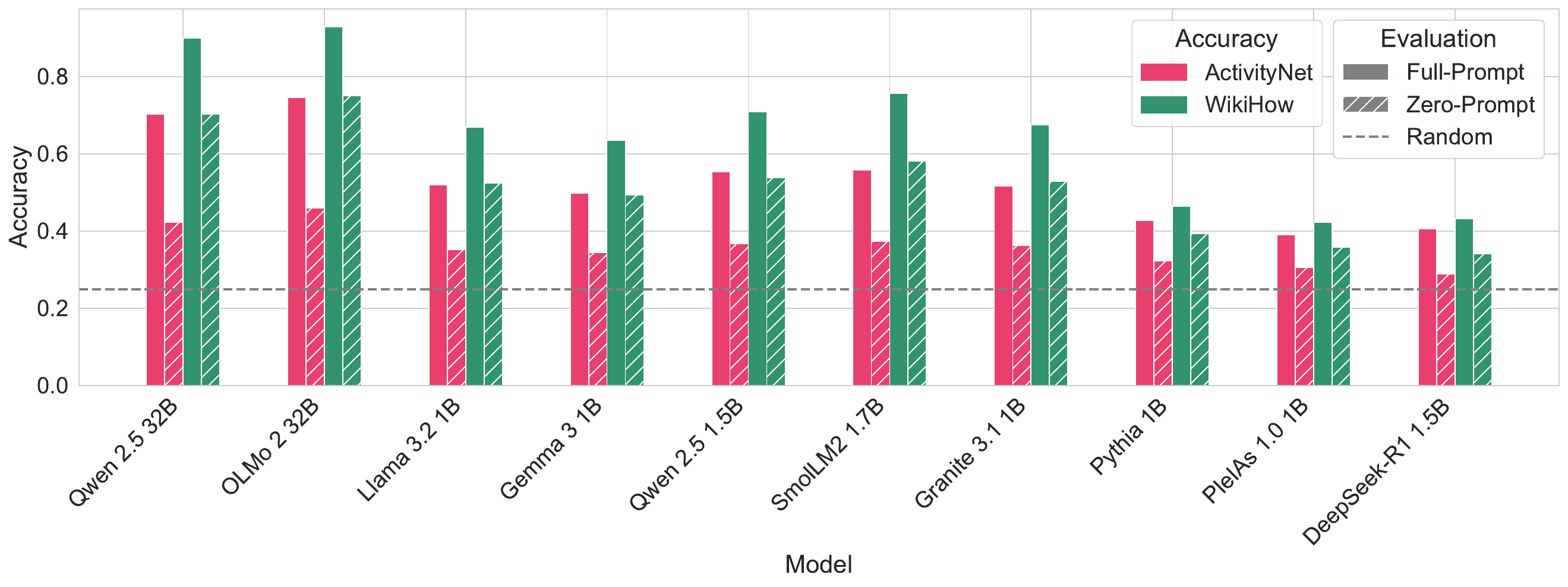}
    \caption{Accuracy of the models evaluated with log-likelihood maximization split by the source of questions.}
    \label{fig:accuracy-source}
\end{figure*}

\begin{table*}[!tbp]
\small
\centering
\begin{tabular}{lcccccc}
\toprule
& \multicolumn{2}{c}{LLM -- A1}
& \multicolumn{2}{c}{LLM -- A2}
& \multicolumn{2}{c}{A1 -- A2} \\
\cmidrule(lr){2-3}
\cmidrule(lr){4-5}
\cmidrule(lr){6-7}
Criterion 
& Agreement & F1
& Agreement & F1
& Agreement & $\alpha$ \\
\midrule
1. Prompt grammar/typos & 0.76 & 0.66 & 0.80 & 0.63 & 0.78 & 0.49 \\
2. Correct option grammar/typos & 0.82 & 0.64 & 0.81 & 0.60 & 0.86 & 0.65 \\
3. Incorrect option(s) grammar/typos & 0.72 & 0.77 & 0.69 & 0.72 & 0.73 & 0.48 \\
4. Incoherent prompt* & 0.86 & 0.12 & 0.87 & 0.13 & 0.88 & 0.30 \\
5. Incoherent correct option* & 0.90 & 0.00 & 0.94 & 0.00 & 0.89 & 0.24 \\
6. Illogical correct option* & 0.92 & 0.33 & 0.91 & 0.00 & 0.88 & -0.05 \\
7. Implausible incorrect option(s) & 0.79 & 0.88 & 0.47 & 0.52 & 0.40 & -0.24 \\
8. Ethical concern* & 0.93 & 0.22 & 0.90 & 0.17 & 0.89 & 0.41 \\
\midrule
9. Best option & 0.89 & --- & 0.95 & --- & 0.86 & --- \\
10. Valid alternatives & 0.03 & --- & 0.03 & --- & 0.07 & ---  \\
11. Worst option & 0.59 & --- & 0.54 & --- & 0.59 & --- \\ 
\bottomrule
\end{tabular}
\caption{Annotation agreement for binary criteria. We show LLM's agreement with every annotator (A1 and A2) along with F1 score, which treats human annotations as ground truth, and inter-annotator agreement together with Krippendorff's $\alpha$. *These categories received small number of positive human annotations (ranging from 3 to 9 records out of 100).}
\label{tab:inter_rater_metrics}
\end{table*}

In Figure~\ref{fig:accuracy-source}, we show the differences in accuracies by question sources. ActivityNet part of the data, which is a suspect for lower-quality questions in HellaSwag, is more challenging to the models. However, taking all our results into account, this complexity is not a compelling benchmark feature, but rather low-quality data that produces unwanted noise in the scores.

Interestingly, the zero-prompt accuracy drop in ActivityNet is also larger than that for the WikiHow. One of the reasons for this might be that WikiHow has generally longer options, from which it is easier to get a general understanding of the situation.

\begin{table*}[!tbp]
\centering
\small
\setlength{\tabcolsep}{4pt}
\begin{tabular}{lcccccccccccc}
\toprule
& \multicolumn{10}{c}{Log-Likelihood} & \multicolumn{2}{c}{Generation} \\
\cmidrule(lr){2-11} \cmidrule(lr){12-13}

& \rotatebox{90}{Llama 3.2 1B}
& \rotatebox{90}{Gemma 3 1B}
& \rotatebox{90}{Qwen 2.5 1.5B}
& \rotatebox{90}{SmolLM2 1.7B}
& \rotatebox{90}{Granite 3.1 1B}
& \rotatebox{90}{Pythia 1B}
& \rotatebox{90}{PleIAs 1B}
& \rotatebox{90}{DeepSeek-R1 1.5B}
& \rotatebox{90}{Qwen 2.5 32B}
& \rotatebox{90}{OLMo 2 32B}
& \rotatebox{90}{Qwen 2.5 32B}
& \rotatebox{90}{OLMo 2 32B} \\
\midrule

Accuracy
& 0.62 & 0.59 & 0.66 & 0.69 & 0.62 & 0.45 & 0.41 & 0.42 & 0.84 & 0.87 & 0.92 & 0.86 \\
\midrule
Prompt grammar/typos
& 0.59 & 0.56 & 0.62 & 0.64 & 0.58 & 0.44 & 0.41 & 0.42 & 0.79 & 0.82 & 0.9 & 0.83 \\
Incoherent prompt
& 0.54 & 0.53 & 0.56 & 0.57 & 0.53 & 0.41 & 0.41 & 0.41 & 0.73 & 0.75 & 0.86 & 0.75 \\
Correct option grammar/typos
& 0.52 & 0.49 & 0.56 & 0.61 & 0.53 & 0.34 & 0.31 & 0.32 & 0.79 & 0.83 & 0.93 & 0.86 \\
Incoherent correct option
& 0.38 & 0.36 & 0.41 & 0.43 & 0.35 & 0.26 & 0.22 & 0.23 & 0.59 & 0.77 & 0.81 & 0.68 \\
Illogical correct option
& 0.47 & 0.44 & 0.48 & 0.5 & 0.45 & 0.36 & 0.33 & 0.33 & 0.67 & 0.75 & 0.67 & 0.59 \\
Incorrect option(s) grammar/typos
& 0.65 & 0.62 & 0.69 & 0.72 & 0.65 & 0.48 & 0.43 & 0.45 & 0.86 & 0.89 & 0.93 & 0.87 \\
Implausible incorrect option(s)
& 0.64 & 0.61 & 0.68 & 0.71 & 0.64 & 0.46 & 0.42 & 0.43 & 0.86 & 0.89 & 0.93 & 0.87 \\

\bottomrule
\end{tabular}
\caption{Performance of all evaluated models on full HellaSwag validation set (top row) and on questions with highlighted issues.}
\label{tab:accuracy_vs_label}
\end{table*}

\section{Human Validation}
\label{sec:app-human}

We collected two sets of human annotations for 100 questions from HellaSwag. The results of human annotators' assessment compared to LLM labels are shown in Table~\ref{tab:inter_rater_metrics}. In most common issues, LLM agrees with at least one of the annotators for the majority of questions, while sometimes the annotators agree with each other less. This might be because of subjectivity of certain issues and difference in expertise. For some criteria, namely 4, 5, 6, and 8, the set of postive labels from humans and LLM was small (less than 10), thus the balanced agreement metrics are low. We investigated separately 20 questions marked with a positive label by the model, and found that at least 60\% of records were labeled correctly for each category (up to 80\% in the ethical concerns).

For the valid alternatives, we computed agreement using intersection over union metric. Both valid alternatives and worst option categories were largely subjective, therefore we did not use them in our further analysis.

\section{Accuracy vs Annotations}
\label{sec:app-accuracy-vs-label}

In Table~\ref{tab:accuracy_vs_label}, we show the performance of all models on the complete benchmark, along with the subsets corresponding to each category of issue annotations. The pattern, where a model scores better on questions with issues in distractors and worse on questions with problematic correct answers or question prompts, is consistent across all models.

\section{Log-Likelihood Evaluation}
\label{app:log-likelihood}

\begin{figure}[!t]
\centering
    \begin{subfigure}[b]{\linewidth}
        \includegraphics[width=\linewidth]{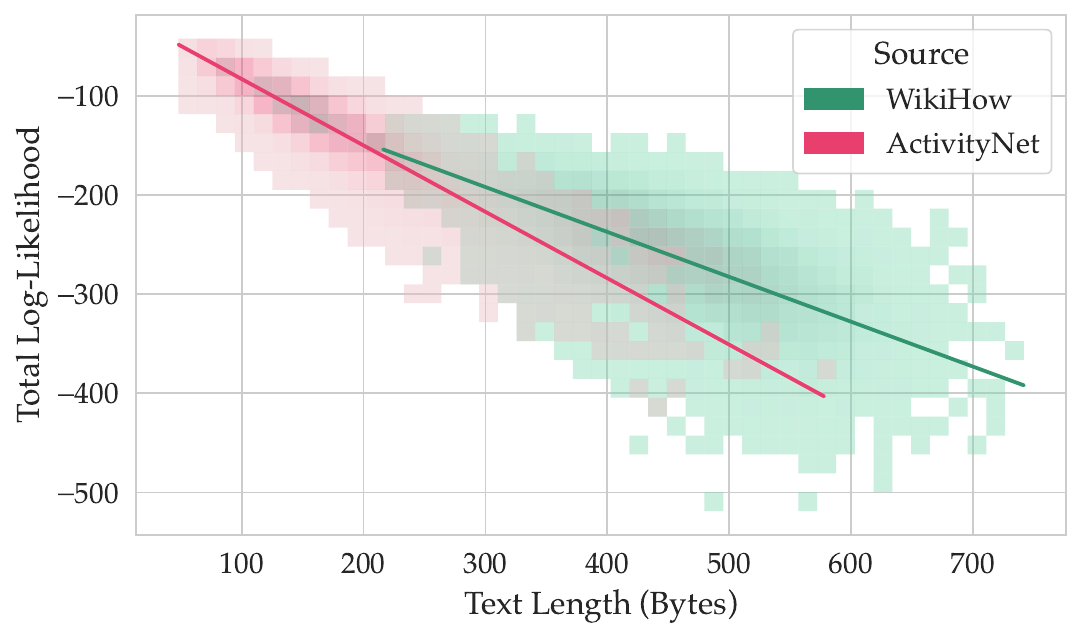}
        \subcaption{Total log-likelihood vs text length in bytes.}
        \label{fig:total_ll}
    \end{subfigure}
    \hfill
    \begin{subfigure}[b]{\linewidth}
        \includegraphics[width=\linewidth]{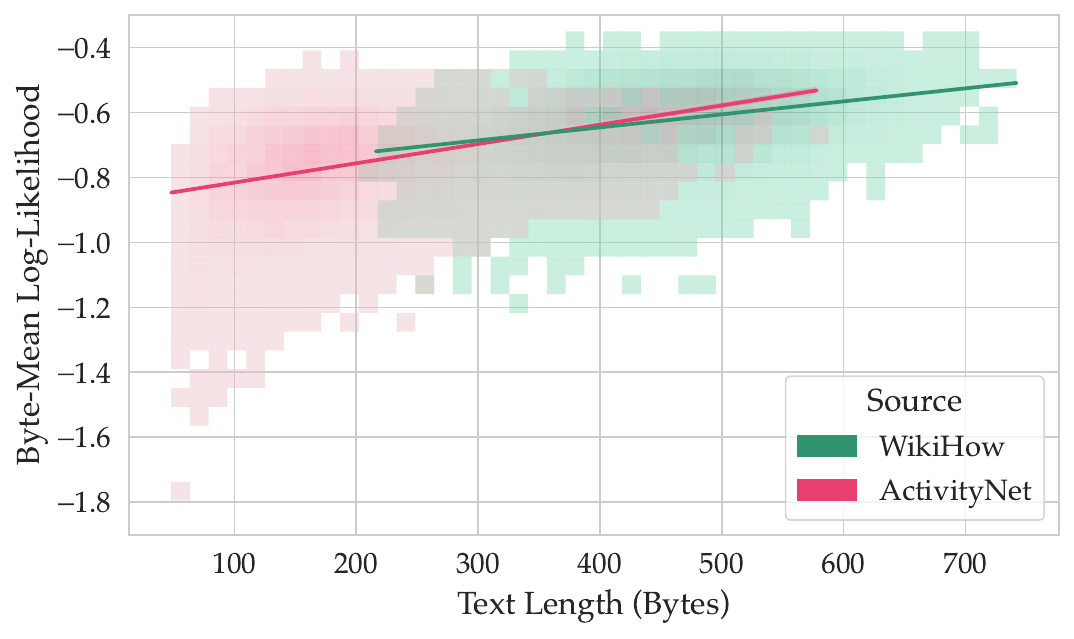}
        \subcaption{Byte-normalized log-likelihood vs text length in bytes.}
        \label{fig:byte_length_ll}
    \end{subfigure}
    \caption{Joint distributions of likelihoods and lengths of the HellaSwag validation set questions for different variants of likelihood computation. The lines represent the trend for each question source.}
    \label{fig:app-lengths}
\end{figure}

We compute the mean log-likelihood using the following formula:
\begin{equation}
    \mathcal{L} = \frac{1}{|\mathcal{S}|} \sum_{t \in \mathcal{S}} \log \mathrm{P}(y_t \mid x_{<t}),
\end{equation}
where $y_t$ is the ground truth token at position $t$, $x_{<t}$ is the preceding sequence of tokens, and $\mathcal{S}$ is the sequence of non-special token positions. The resulting value will have non-positive values, and the higher it is, the more plausible the option is, according to the model. 

We also compare the correlation with answer lengths for two other formulations of log-likelihood. In Figure~\ref{fig:app-lengths}, we show joint distributions for \textbf{total} log-likelihood:
\begin{equation}
    \mathcal{L}_{t} = \sum_{t \in \mathcal{V}} \log \mathrm{P}(y_t \mid x_{<t}),
\end{equation}
and \textbf{byte-normalized} log-likelihood:
\begin{equation}
    \mathcal{L}_{b} = \frac{1}{\mathrm{B}}\sum_{t \in \mathcal{V}} \log \mathrm{P}(y_t \mid x_{<t}),
\end{equation}
where $y_t$ is the ground truth token at position $t$, $x_{<t}$ is the preceding sequence of tokens, $\mathrm{B}$ is the sequence length in bytes, and $\mathcal{V}$ is the set of valid (non-special) token positions. As with the mean log-likelihood, there is a visible correlation between answer length and log-likelihood in both cases (negative, in Figure~\ref{fig:total_ll}).

\section{Grammar Issues}
\label{app:grammar-issues}

\begin{figure*}[!t]
    \centering
    \includegraphics[width=\linewidth]{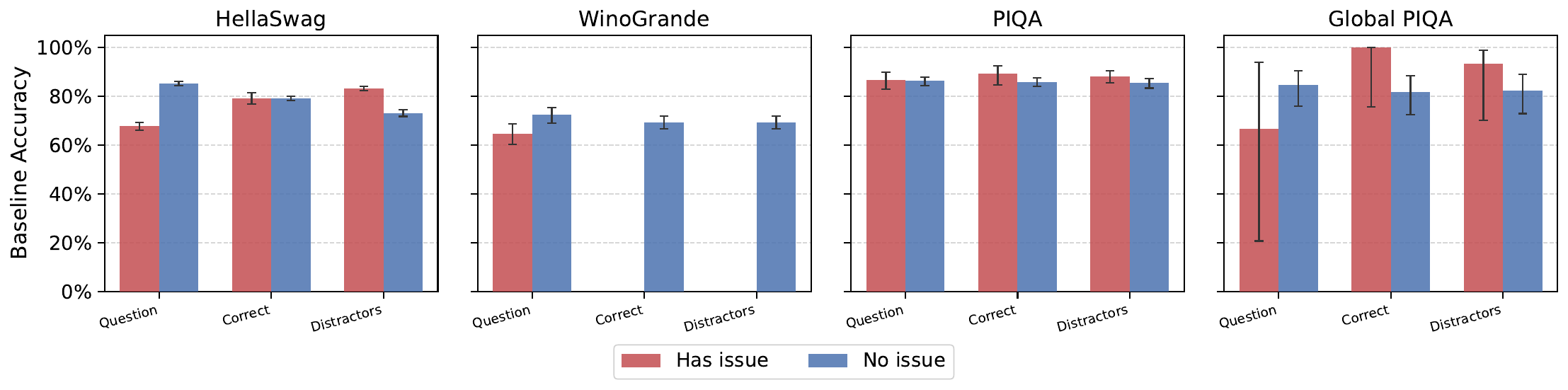}
    \caption{Accuracy of Gemma 3 12B~\citep{google2024gemma3} on four commonsense reasoning benchmarks, split by the presence of grammar issues.}
    \label{fig:grammar-accuracy}
\end{figure*}

\begin{figure*}
    \centering
    \includegraphics[width=\linewidth]{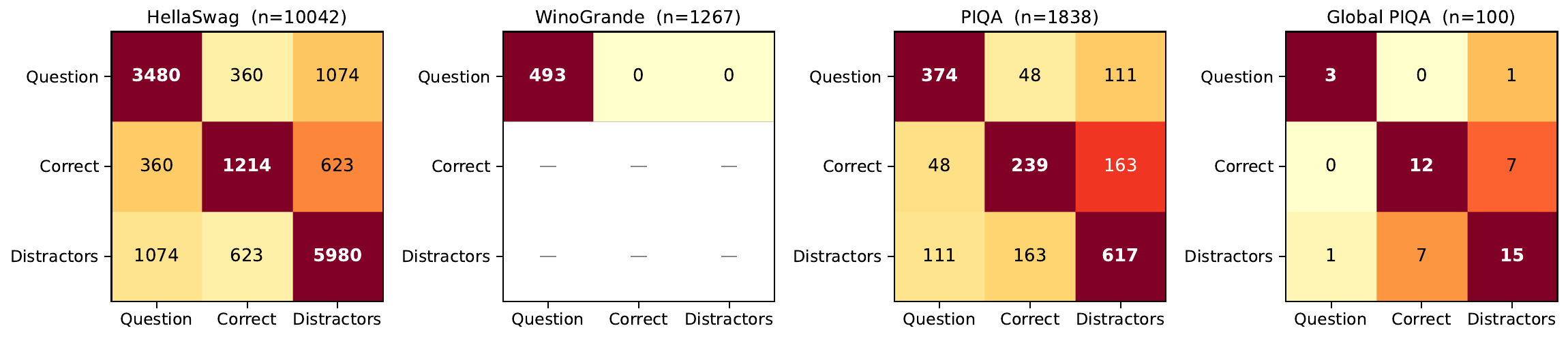}
    \caption{Grammatical issue overlap in commonsense reasoning benchmarks. Grammar issues are evaluated via guided decoding from Gemma 4 E4B~\citep{gemma4_hf_blog_2026}.}
    \label{fig:grammar-overlap}
\end{figure*}

The observations made in Section~\ref{sec:issues-accuracy} for HellaSwag are present in other commonsense reasoning benchmarks as well (see Figure~\ref{fig:grammar-accuracy}). Winogrande has no issues with answer options, whereas the issues in the question prompt make the task harder. In PIQA and Global PIQA, grammar issues in the distractors make the task easier, with a similar pattern in the golden answers. A potential reason is that there is overlap between these issues in the benchmark items (see Figure~\ref{fig:grammar-overlap}). The issues in the question prompts in Global PIQA also decrease the model's accuracy, yet this part has a large error bar, having only three questions as a support set.

\section{Near-duplicate Contamination}
\label{app:dist-to-train}

\begin{figure}
    \centering
    \includegraphics[width=\linewidth]{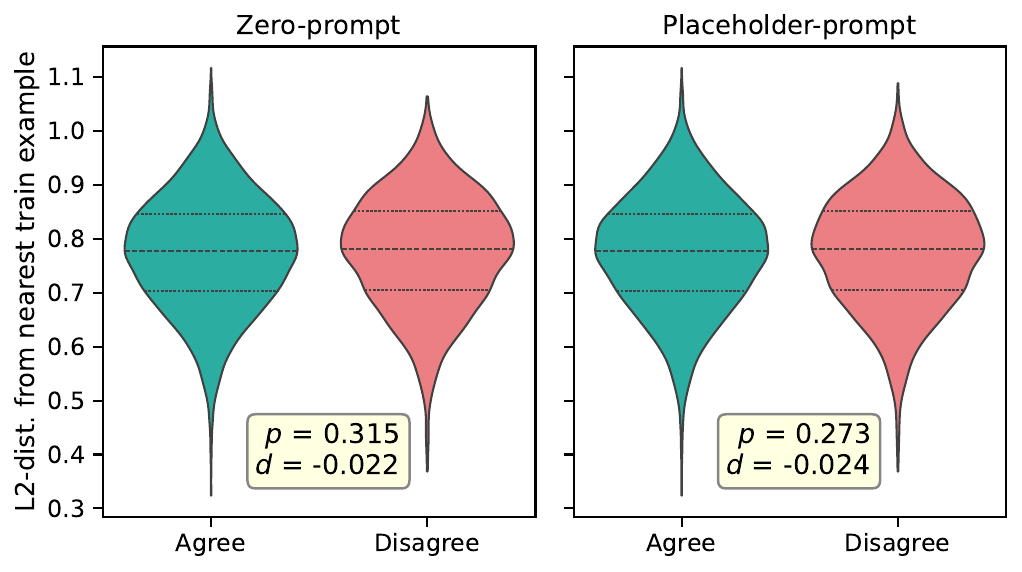}
    \caption{Distribution of L2 distances to near-duplicates in the train set for zero-/placeholder-prompt agreement for Gemma 3 1B. For each pair, we present the p-value of the t-test and Cohen's d.}
    \label{fig:dist-to-train-gemma}
\end{figure}

\begin{figure}
    \centering
    \includegraphics[width=\linewidth]{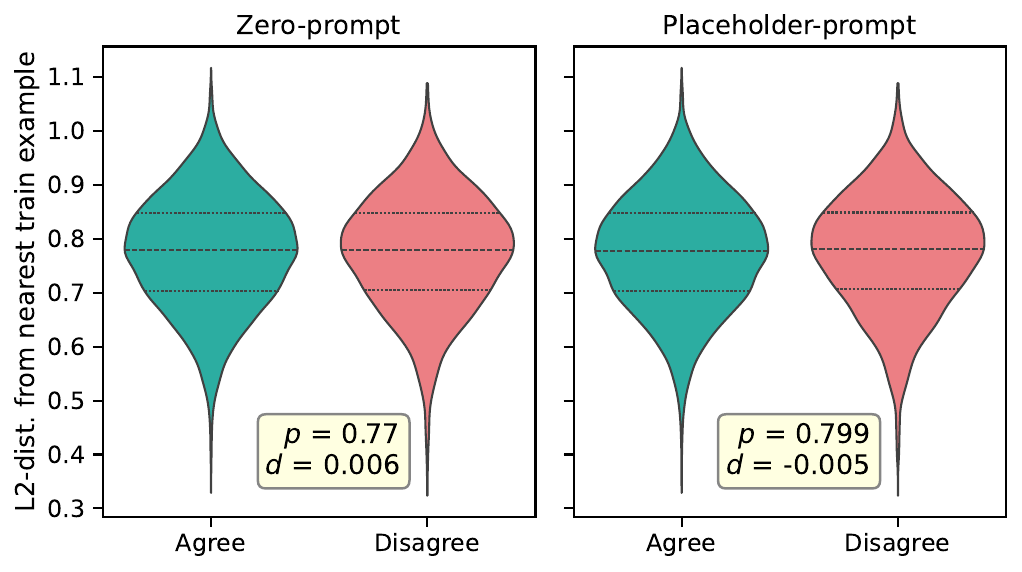}
    \caption{Distribution of L2 distances to near-duplicates in the train set for zero-/placeholder-prompt agreement for Qwen2.5 1.5B. For each pair, we present the p-value of the t-test and Cohen's d.}
    \label{fig:dist-to-train-qwen}
\end{figure}

\begin{figure}
    \centering
    \includegraphics[width=\linewidth]{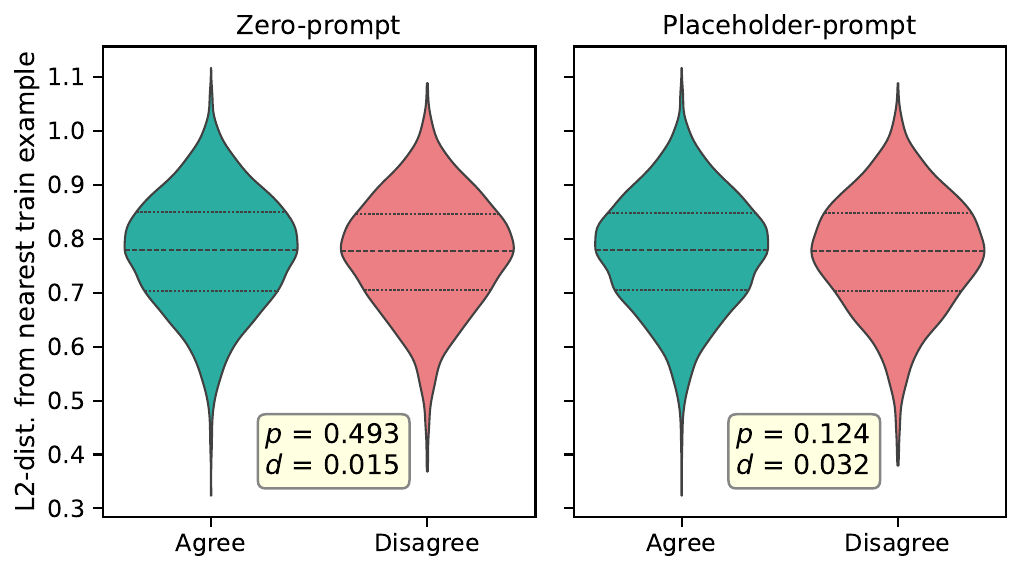}
    \caption{Distribution of L2 distances to near-duplicates in the train set for zero-/placeholder-prompt agreement for SmolLM2 1.7B. For each pair, we present the p-value of the t-test and Cohen's d.}
    \label{fig:dist-to-train-smollm}
\end{figure}

\begin{figure}
    \centering
    \includegraphics[width=\linewidth]{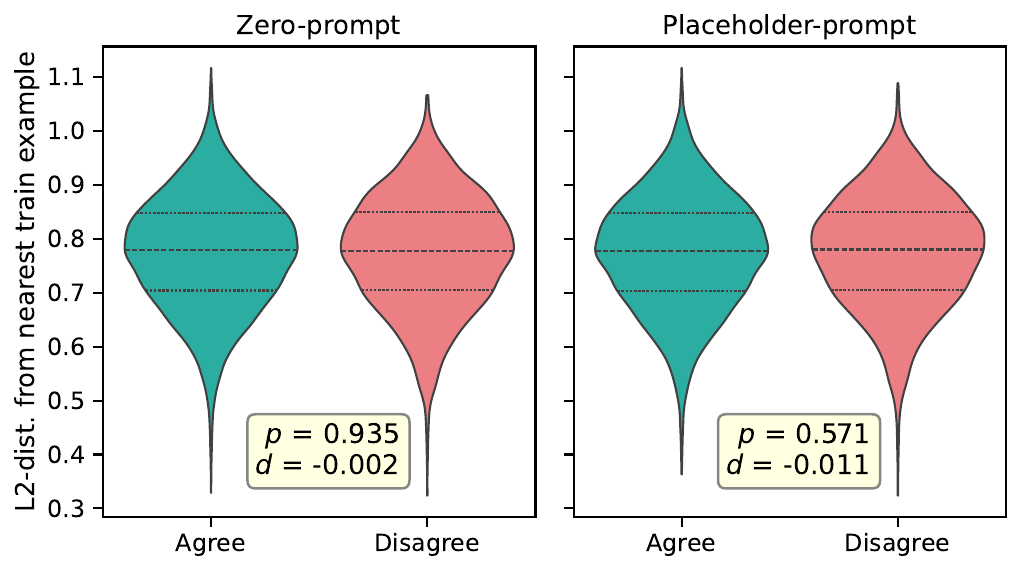}
    \caption{Distribution of L2 distances to near-duplicates in the train set for zero-/placeholder-prompt agreement for PleIAs 1.0 1B. For each pair, we present the p-value of the t-test and Cohen's d.}
    \label{fig:dist-to-train-pleias}
\end{figure}

In Figures~\ref{fig:dist-to-train-gemma}, \ref{fig:dist-to-train-qwen}, \ref{fig:dist-to-train-smollm}, and \ref{fig:dist-to-train-pleias}, we show additional example of L2 distances distribution between train and validation sets in HellaSwag. The same pattern of no significant difference is showing in the models with both closed data and open data.

\section{Annotations}
\label{sec:app-annotation}

In Figure~\ref{fig:prompt-1}, we report the prompt we used for annotations with Gemini 2.5 Flash.

\section{LLM Usage Statement}

In this work, we utilized AI tools for paraphrasing and grammar correction (Grammarly) and assistance with data visualization design (Gemini, ChatGPT, Claude Code).

\begin{figure*}[t]
\small
\centering
\begin{verbatim}
You are an expert linguist and professional data annotator. Your task is to analyze a multiple-choice question 
from the "HellaSwag" commonsense reasoning benchmark. You must output your analysis in strict JSON format.

## Context
Here is the dataset entry to analyze:

**Question Context:** {data['query']}

**Answer Options:**
{options_text}

**The Intended Correct Answer Option:** {correct_letter}

## Instructions
Analyze the validity of this question based on the criteria below. 

### Criteria Definitions
1. **prompt_grammar_error**: (1/0) The question context has grammatical errors or typos.
2. **correct_grammar_error**: (1/0) The intended correct answer text has grammatical errors or typos.
3. **incorrect_grammar_error**: (1/0) At least one incorrect answer option has grammatical errors or typos.
4. **prompt_incoherent**: (1/0) The question context is incoherent or nonsensical.
5. **correct_incoherent**: (1/0) The intended correct answer text is incoherent or nonsensical.
6. **correct_illogical**: (1/0) The intended correct answer makes no logical sense *specifically in the context 
of the question*.
7. **incorrect_implausible**: (1/0) At least one incorrect answer option is *fundamentally incoherent, 
structurally nonsensical, or contains clear evidence of hallucinated or garbled text (e.g., random word sequences, 
impossible object names, or distinct incoherent phrases or constructions the sense of which is hard to establish)*. 
This assessment is based on the option's internal validity, independent of the question's context.
8. **ethical_concern**: (1/0) At least one component of the question (context or any option) raises serious ethical 
concerns (e.g., promoting hate speech, illegal acts, or harmful stereotypes).

### Qualitative Analysis
9. **best_option**: Which option (A/B/C/D) is the absolute best fit?
10. **valid_alternatives**: List other options (A/B/C/D) that fit equally well. If none, strictly return "None".
11. **worst_option**: Which option fits the prompt the least?

## Output Format
You must respond with a single valid JSON object. Do not include markdown code blocks. For example:

{
  "prompt\_grammar\_error": 1,
  "correct\_grammar\_error": 0,
  "incorrect\_grammar\_error": 1,
  "prompt\_incoherent": 0,
  "correct\_incoherent": 0,
  "correct\_illogical": 0,
  "incorrect\_implausible": 1,
  "ethical\_concern": 0,
  "best\_option": "A",
  "valid\_alternatives": ["C", "D"],
  "worst\_option": "B"
}
\end{verbatim}
\caption{Prompt for LLM annotations.}
\label{fig:prompt-1}
\end{figure*}

\end{document}